%% file: main.tex
\documentclass{article} 
\pdfoutput=1
\usepackage{aloe2022,times}

\input{math_commands.tex}

\usepackage{hyperref}
\usepackage{url}
\usepackage{natbib}
\usepackage{microtype}
\usepackage{graphicx}
\usepackage{subfigure}
\usepackage{booktabs} 
\usepackage{amsmath}
\usepackage{bbding}

\usepackage{algorithm}
\usepackage{algorithmic}
\usepackage{todonotes}
\usepackage{bbm}

\title{When to Go, and When to Explore: \\ The Benefit of Post-Exploration in Intrinsic Motivation}

\iclrfinalcopy
\author{Zhao Yang, Thomas M. Moerland, Mike Preuss and Aske Plaat \\
Leiden Institute of Advanced Computer Science \\
Leiden University, The Netherlands \\
}
\begin{document}

\maketitle

\begin{abstract}
Go-Explore achieved breakthrough performance on challenging reinforcement learning (RL) tasks with sparse rewards. The key insight of Go-Explore was that successful exploration requires an agent to first return to an interesting state (`Go'), and only then explore into unknown terrain (`Explore'). We refer to such exploration after a goal is reached as `post-exploration'. In this paper we present a systematic study of post-exploration, answering open questions that the Go-Explore paper did not answer yet. First, we study the isolated potential of post-exploration, by turning it on and off within the same algorithm. Subsequently, we introduce new methodology to adaptively decide {\it when} to post-explore and for {\it how long} to post-explore. Experiments on a range of MiniGrid environments show that post-exploration indeed boosts performance (with a bigger impact than tuning regular exploration parameters), and this effect is further enhanced by adaptively deciding when and for how long to post-explore. In short, our work identifies {\it adaptive} post-exploration as a promising direction for RL exploration research. 
\end{abstract}

\section{Introduction} \label{introduction}
\begin{figure}[t]
    \centering
    \includegraphics[scale=0.35]{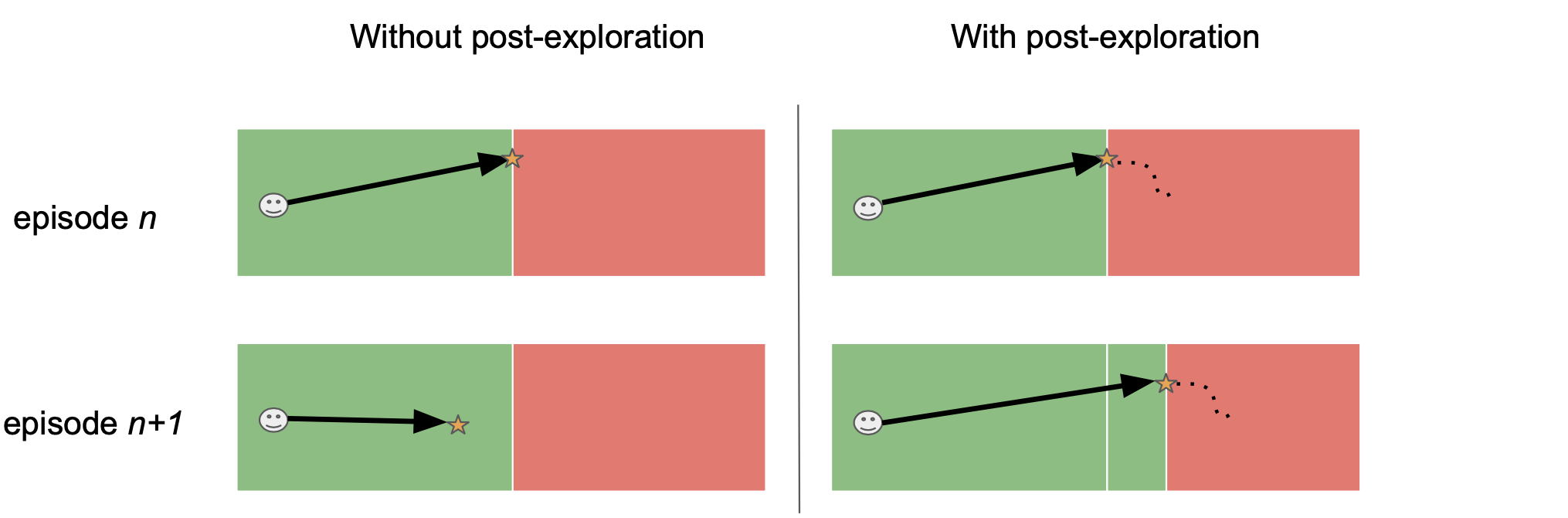}
    \caption{Conceptual illustration of post-exploration. Each box displays the entire state space, where green and red denote the (currently) explored and unexplored regions, respectively. {\bf Left}: Goal-based exploration without post-exploration. The top graph shows the agent reaching a current goal, after which the episode is terminated (or a next goal in the green area is chosen). Therefore, at the next episode (bottom) the agent will again explore within the known (green) region), often leaving the unknown (red) area untouched. {\bf Right}: Goal-based exploration with post-exploration. The top figure shows the agent reaching a particular goal, from which the agent now post-explores for several steps (dashed lines). Thereby, the known area (green) is pushed outwards. On a next episode, the agent may now also select a goal in the expanded region, gradually expanding its knowledge of the reachable state space. }
    \label{fig:intuition}
\end{figure}

Go-Explore~\citep{ecoffet2021first} achieved breakthrough performance on challenging reinforcement learning (RL) tasks with sparse rewards, most notably achieving state-of-the-art, `super-human' performance on Montezuma's Revenge, a long-standing challenge in the field. The key insight behind Go-Explore is that proper exploration should be structured in two phases: an agent should first attempt to get back to a previously visited, interesting state (`Go'), and only then explore into new, unknown terrain (`Explore'). Thereby, the agent gradually expands its knowledge base, an approach that is visualized in Fig.\ref{fig:intuition}. We propose to call such exploration after the agent reached a goal {\it post-exploration} (to contrast it with standard exploration).

There are actually two variants of Go-Explore in the original paper: one in which we directly reset the agent to an interesting goal ({\it restore-based} Go-Explore), and one in which the agent has to act to get back to the proposed goal ({\it policy-based} Go-Explore). In this work, we focus on the latter approach, which is technically part of the literature on intrinsic motivation, in particular intrinsically motivated goal exploration processes (IMGEP)~\citep{colas2020intrinsically}. Note that post-exploration does not require any changes to the IMGEP framework itself, and can therefore be easily integrated into other existing work in this direction. 

While Go-Explore gave a strong indication of the potential of post-exploration, it did not structurally investigate the benefit and possible extensions of the approach. First of all, Go-Explore was compare to other baseline algorithms, but post-exploration itself was never switched on and off in the same algorithm. Thereby, the isolated performance gain of post-exploration remains unclear. Second, it also remains unclear {\it when} we should post-explore (Go-Explore does this at every trial) and {\it for how long} we should post-explore (Go-Explore does this for a fixed number of steps). 

Therefore, the present paper studies {\it adaptive} post-exploration. In particular, we make the decision to post-explore a function of the {\it novelty} of the reached goal, and the depth of post-exploration a function of the length of the goal-reaching part of the episode. Experiments in a range of MiniGrid tasks show that post-exploration provides a strong isolated benefit over standard IMGEP algorithms, which can be further enhanced by adaptively deciding when and for how long to post-explore. As a smaller contribution, we also cast Go-Explore into the IMGEP framework, and show how it can be combined with hindsight experience replay (HER) ~\citep{NIPS2017_453fadbd} to make more efficient use of the observed data. In short, our work presents a systematic study of post-exploration, and identifies {\it adaptive} post-exploration as a promising direction for future RL exploration research. 

\section{Related Work}
\label{related}
Go-Explore is a variant of a intrinsically motivated goal exploration processes (IMGEP) (surveyed by ~\citet{colas2020intrinsically}) which generally consists of three phases: 1) defining/learning a goal space, 2) sampling an interesting goal from this space, and 3) getting to the sampled goal based on information from previous episodes. Regarding the first step, Go-Explore~\citep{ecoffet2021first} uses a predefined goal space based on a downscaling of the state space. Other IMGEP approaches, like Goal-GAN~\citep{pmlr-v80-florensa18a} use a generative adversarial network to learn a goal space of appropriate difficulty. As an alternative, AMIGo~\citep{campero2021learning} trains a teacher to propose goals for a student to reach. In this work, we largely follow the ideas of Go-Explore, and use a predefined goal space, since our research questions are directed at the effects of post-exploration.

After a goal space is determined, we next need a sampling strategy to set the next goal. Example strategies to select the next goal include novelty~\citep{ecoffet2021first,pitis2020maximum}, learning progress~\citep{colas2019curious,portelas2020teacher},  diversity~\citep{pong2019skew,warde-farley2018unsupervised}, and uniform sampling ~\citep{eysenbach2018}. In this work, we follow the last approach, since it 1) gave good performance and 2) puts pressure on the agent to adaptively decide when to post-explore. 

In the third IMGEP phase, we need to get back to the proposed goal. One variant of Go-Explore simply resets the agent to the desired goal, but this requires an environment which can be set to an arbitrary state. Instead, we follow the more generic `policy-based Go-Explore' approach, which uses a goal-conditioned policy network to get back to a goal. The concept of goal-conditioning, which was introduced by~\citet{schaul2015universal}, is also the common approach in IMGEP approaches. A well-known addition to goal-based RL approaches is hindsight experience replay~\citep{NIPS2017_453fadbd}, which allows the agent to make more efficient use of the data through counterfactual goal relabelling. Although Go-Explore did not use hindsight in their work, we do include it as an extension in our approach. 

After we manage to reach a goal, most IMGEP literature samples a new goal and either resets the episode or continues from the reached state. The main contribution of Go-Explore was the introduction of post-exploration, in which case we temporarily postpone selection of a new goal (Tab.~\ref{tab:overview}). While post-exploration is a new phenomenon in reinforcement learning literature, it is a common feature of most planning algorithms, where we for example select a particular node on the frontier, go there first, and then unfold the search tree into unknown terrain. 

\begin{table}[!t]
    \centering
    \caption{Overview of moment of exploration in IMGEP papers. Most IMGEP approaches, like Goal-GAN~\citep{pmlr-v80-florensa18a}, CURIOUS~\citep{colas2019curious}, DIAYN~\citep{pong2019skew}, and AMIGo~\citep{campero2021learning}, only explore during goal reaching. Restore-based Go-Explore, which directly resets to a goal, only explore after the goal is reached. Our work, similar to policy-based Go-explore, explores both during goal-reaching and after goal reaching.}
    \begin{tabular}{c|c|c}
        \textbf{Approach} & \textbf{Exploration during goal reaching} & \textbf{Post-exploration} \\
        \hline
        IMGEP & \Checkmark & \XSolidBrush \\
        Restore-based Go-Explore & \XSolidBrush & \Checkmark \\
        Policy-based Go-Explore + this paper & \Checkmark & \Checkmark \\
    \end{tabular}

    \label{tab:overview}
\end{table}

\section{Background}
We adopt a Markov Decision Processes (MDPs) formulation defined as the tuple $M=(\mathcal{S}, \mathcal{A}, P, R, \gamma)$ ~\citep{sutton2018reinforcement}. Here, $\mathcal{S}$ is a set of states, $\mathcal{A}$ is a set of actions the agent can take, $P$ specifies the transition function of the environment, and $R$ is the reward function. At timestep $t$, the agent observes state $s_t \in \mathcal{S}$, selects an action $a_t \in \mathcal{A}$, after which the environment returns a next state $s_{t+1} \sim P(\cdot|s_t,a_t)$ and associated reward $r_t = R(s_t,a_t,s_{t+1})$. We act in the MDP according to a policy $\pi(a|s)$, which maps a state to a distribution over actions. When we unroll the policy and environment for $T$ steps, define the cumulative reward (return) of the resulting trace as  $\sum ^{T}_{k=0} \gamma^k \cdot r_{t+k+1}$, where $\gamma \in [0,1]$ denotes a discount factor. Define the state-action value $Q^\pi(s,a)$ as the {\it expected} cumulative reward under some policy $\pi$ when we start from state $s$ and action $a$, i.e.,

\begin{equation}
    Q^\pi(s,a) = \mathbb{E}_{\pi,P} \Big[\sum ^{T}_{k=0} \gamma^k \cdot r_{t+k}|s_t=s,a_t=a \Big]
\end{equation} 

Our goal is to find the {\it optimal} state-action value function $Q^*(s,a)=\max_{\pi} Q^\pi(s,a)$, from which we may at each state derive the optimal policy $\pi^*(s)=\argmax_a Q^*(s,a)$. A well-known RL approach to solve this problem is {\it Q-learning} \citep{watkins1992q}. Tabular Q-learning maintains a table of Q-value estimates $\hat{Q}(s,a)$, collects transition tuples $(s_t,a_t,r_t,s_{t+1})$, and subsequently updates the tabular estimates according to 
\begin{equation}
    \hat{Q}(s_t,a_t) \gets \hat{Q}(s_t,a_t) + \alpha \cdot[r_{t}+\gamma \cdot \max_{a} \hat{Q}(s_{t+1},a) - \hat{Q}(s_t,a_t)]
    \label{eq:q}
\end{equation} where $\alpha \in [0,1]$ denotes a learning rate. Under a policy that is greedy in the limit with infinite exploration (GLIE) this algorithm converges on the optimal value function \citep{watkins1992q}. 

\section{Methods}
We will first describe how we cast Go-Explore into the general IMGEP framework (Sec. \ref{sec_imgep}), and subsequently introduce our new methodology to adaptively decide when and for how long to post-explore (Sec. \ref{sec_post_exploration}).

\subsection{IMGEP} \label{sec_imgep}
Our work is based on the IMGEP framework shown in Appendix~\ref{alg}, Alg~\ref{alg:pe}. Since this paper attempts to study the fundamental benefit of post-exploration, we try to simplify the problem setting as much as possible, to avoid interference with other issues. We therefore choose to study tabular RL problems, in which we define the goal space as the set of states we have observed so far. The goal space $G$ is initialized by executing a random policy for one episode, while new states in future episodes augment the set. 

For goal sampling we take uniform samples from the goal space, as for example also used by ~\citet{eysenbach2018}. This approach empirically provided best performance, and also increases the impact of the decision when to post-explore. Note that we could also use other methods, both for defining the goal space and for sampling new goals (for example based on novelty or learning progress), but these choices allow us to focus on the benefit of post-exploration. 

To get (back) to a selected goal, we train a tabular goal-conditioned Q-learning agent. The update rule of the value function is similar to the standard  Q-learning update shown in Eq.~\ref{eq:q}, except that $Q$ is conditioned on the selected goal $g$, i.e., $Q(s,a,g)$. Agents are trained on the goal-conditioned reward function $R_g$, which is a one-hot indicator that fires when the agent manages to reach the specified goal 

\begin{equation}
    R_g(s,a,s') =  \mathbbm{1}_{s'=g},
\end{equation} 

Note that different goal-conditioned rewards functions, different back-up strategies and/or the use of function approximation are of course possible as well. 

\subsection{Post-exploration} \label{sec_post_exploration}
\label{PE}
The general concept of post-exploration was already introduced in Figure~\ref{fig:intuition}. The main benefit of this approach is that it increases our chance to step into new, unknown terrain. Based on this intuition, we will now discuss new methodology to adaptively decide {\it when} to post-explore and for {\it how long} to post explore. 

\paragraph{When to post-explore} As is directly visible from Figure~\ref{fig:intuition}, post-exploration is likely most beneficial when performed from a state at the border of our currently explored region. After all, when we perform post-exploration from a state in the middle of the known region, it will likely only waste resources (and not discover anything new), while post-exploration from the border does have much potential. We propose to use the {\it novelty} of the particular state as a measure of its location in the explored region: when a state is only visited a few times, then it is more likely to reside at the border of our explored region (although other measures, like a connectivity graph of the visited states, is also possible). 

We therefore propose to make the probability to post-explore in a particular reached goal $g$ a function of the number of times we have visited that particular goal. We therefore track the number of times we visited a particular goal $n(g)$, and specify the probability to post-explore ({\it when} to post-explore) as 

\begin{equation}
    p_{pe}(g)=\left (\frac{1}{n(g)} \right) ^{\beta}, \beta \in [0,\infty). 
    \label{eq:beta}
\end{equation} 

Here, $\beta$ is a temperature parameter that allow us to scale the amount of post-exploration. When $\beta$ is 0, the agent will always post-explore after reaching the goal, while $\beta \to \infty$ ensures the agent will never post-explore. 

\paragraph{How long to post-explore} We will denote the number of steps an agent post-explores after reaching a goal as $n_{pe}$. While Go-Explore always explored for a fixed number of steps ($n_{pe}$ fixed), we hypothesize that the duration of post-exploration may also influence performance. Intuitively, the longer the path the agent took to reach the given goal, the more time it may want to spend in post-exploration (since it is expensive to get back there). We therefore propose to let the agent post-explore for a percentage $p_{pe}$ of the length of the whole trajectory:

\begin{equation}
    n_{pe} = p_{pe} \cdot n_{ep}, \label{eq_n_pe}
\end{equation} 

where $n_{ep}$ is the length of the goal-reaching part of the episode, and $p_{pe}$ is a hyperparameter of our algorithm. 

\paragraph{Hindsight relabelling}
The agent in principle only observes a non-zero reward when it successfully reaches the goal, which may not happen at every attempt. We may improve sample efficiency (make more efficient use of each observed trajectory) through {\it hindsight relabelling}~\citep{NIPS2017_453fadbd}. With hindsight, we imagine states in our observed trajectory were the goals we actually attempted to reach, which allows us to extract additional information from them (`if my goal had been to get to this state, then the previous action sequence would have been a successful attempt'). Pseudocode for hinsight relabelling is provided in Appendix~\ref{alg}, Alg.~\ref{alg:hr}. In particular, we choose to always relabel 50\% of the entire trajectory (goal reaching plus post-exploration),  i.e., half of the states in each trajectory are imagined as if they were the goal of that episode. We choose to always relabel the full post-exploration part, because it likely contains the most interesting information, and randomly sample the remaining relabelling budget from the part of the episode before the goal was reached. 

\begin{figure}[!tt]
    \centering
    \includegraphics[scale=0.5]{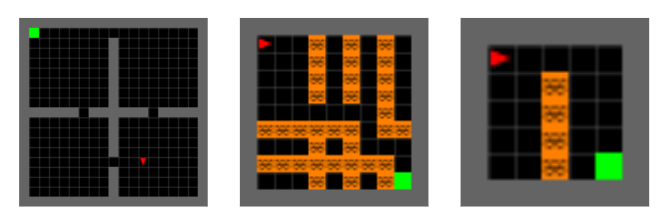}
    \caption{FourRooms, LavaCrossing and LavaGap. The latter two environments are procedurally generated based on environment seeds.}
    \label{fig:envs}
\end{figure}

\section{Experimental Setup}
We test our work on three MiniGrid environments ~\citep{gym_minigrid}, visualized in Fig.~\ref{fig:envs}. The \texttt{MiniGrid-FourRooms-v0} only has a single lay-out, while the \texttt{MiniGrid-LavaCrossingS11N5-v0} and \texttt{MiniGrid-LavaGapS7-v0} environments are procedurally generated. Results on these latter two environments are averaged over 10 seeds, i.e., 10 independently drawn instances of the task. For evaluation, we test the ability of the agent to reach every possible state in the state space. This checks to what extent the goal-conditioned value function is able to reach a given goal, when we execute the greedy policy (turn exploration off). All our results report the total number of environment steps on the x-axis. Therefore, since an episode with post-exploration takes longer, it will also contribute more steps to the x-axis (i.e., we report performance against the total number of unique environment calls). Curves display mean performance over time, including the standard error over 5 repetitions for each experiment. For all experiments, we set $\epsilon=0.1$ (exploration during goal reaching), $\epsilon_{pe}=1.0$ (full exploration during post-exploration), $\beta=0$, and $p_{pe}=0.5$, unless otherwise noted. Full details on hyperparameters can be found in Appendix.~\ref{para}.

\section{Results}
\label{results}
We split our results up into five research questions: 1) does post-exploration work in general, 2) to what extent should the agent also explore during goal reaching, 3) when should we post-explore, 4) for how long should we post-explore, and 5) is post-exploring also beneficial in continuing tasks, where we sample a next goal without resetting the agent to an initial state. Each research questions will be discussed below. 

\begin{figure}[!t]
   \begin{minipage}{0.32\textwidth}
     \centering
     \includegraphics[width=.8\linewidth]{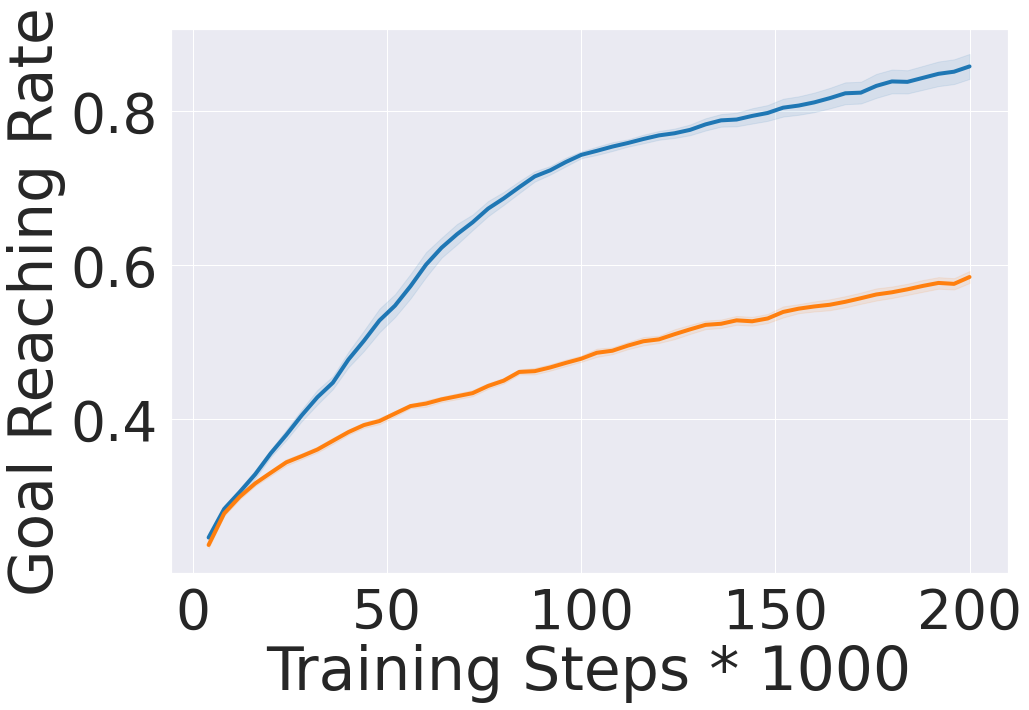}
   \end{minipage}\hfill
   \begin{minipage}{0.32\textwidth}
     \centering
     \includegraphics[width=.82\linewidth]{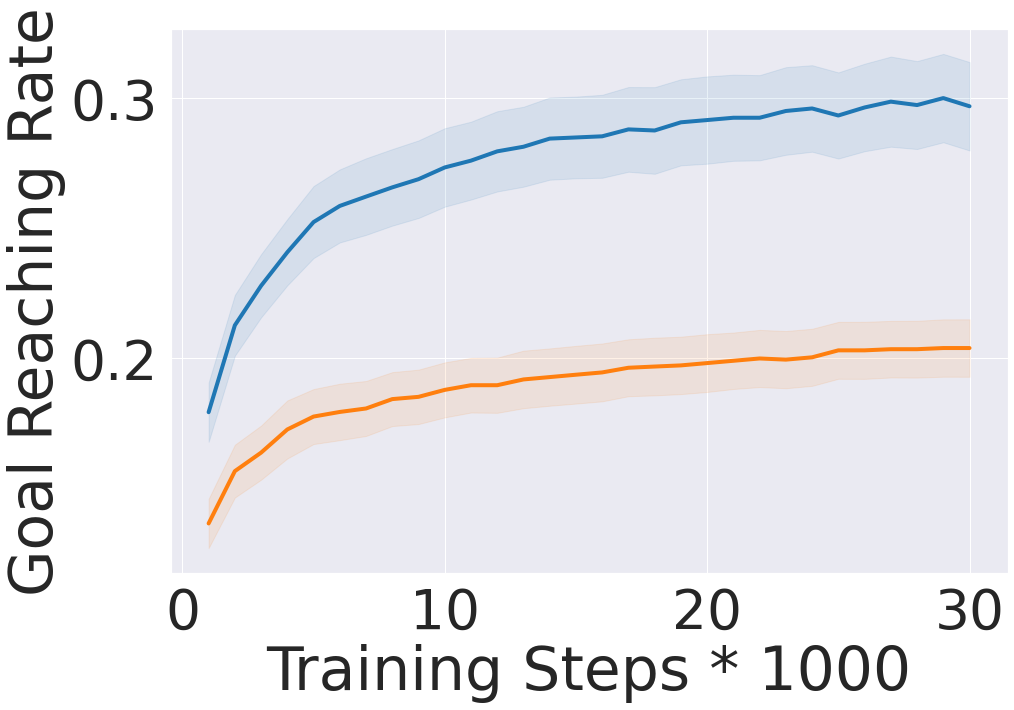}
   \end{minipage}
   \begin{minipage}{0.32\textwidth}
     \centering
     \includegraphics[width=.82\linewidth]{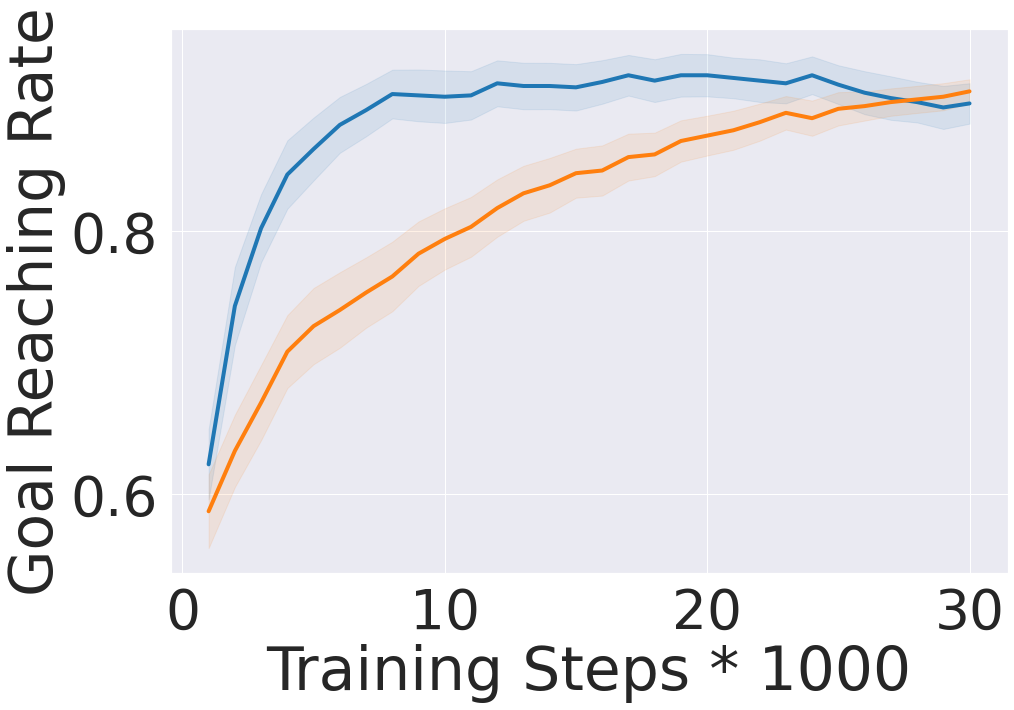}
   \end{minipage}
   \caption{IMGEP agents with (blue) and without (orange) post-exploration. {\bf Left}: Performance on FourRooms, with results average over three different values of $\epsilon$ ($0,0.1,0.3$). {\bf Middle}: Performance on LavaCrossing, averaged over 10 different environment seeds. {\bf Right}: Performance in LavaGap, averaged over 10 different environment seeds. Overall, agents with post-exploration outperform agents without post-exploration.}
   \label{fig: PEworks}
\end{figure}

\subsection{Does post-exploration work?}
Fig.~\ref{fig: PEworks} compares an IMGEP agent with post-exploration to an IMGEP agent without exploration in the three test environments. On all three environments the agent with post-exploration outperforms the baseline agent significantly (on its average ability to reach an arbitrary goal in the state space). 

\begin{figure}[!t]
    \centering
    \includegraphics[scale=0.33]{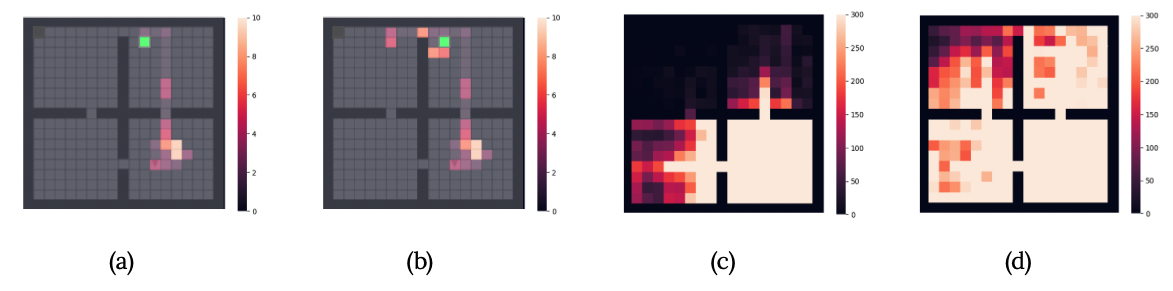}
    \caption{Comparison of characteristic traces (a,b) and coverage (c,d) for agents without post-exploration (a,c) and with post-exploration (b,d). Colour bars indicate the number of visitations, green square indicates the selected goal in a particular episode. (a): Standard exploration towards goal without post-exploration. (b): With post-exploration the agent manages to reach the next room. (c): Coverage after 200k training steps without post-exploration. (d): Coverage after 200k training steps with post-exploration. The boundary of the coverage with post-exploration clearly lies further ahead.}
    \label{fig:peNcover}
\end{figure}

A graphical illustration of the effect of post-exploration is provided in Fig.~\ref{fig:peNcover}. Part a shows a characteristic trace for an agent attempting to reach the green square {\it without} post-exploration, while b shows a trace for an agent {\it with} post-exploration. In a, we see that the agent is able to reach the goal square, but exploration subsequently stops, and the agent did not learn anything new. In part b, the agent with post-exploration subsequently manages to enter the next room. This graph therefore gives a practical illustration of our intuition from Fig.~\ref{fig:intuition}.  

To further illustrate this effect, Fig.~\ref{fig:peNcover} c and d show a visitation heat map for the agent after 200k training steps, without post exploration (c) and with post-exploration (d). The agent with post-exploration (c) has primarily explored goals around the start region (in the bottom-right chamber). The two rooms next to the starting room have also been visited, but the agent barely managed to get into the top-left room. In contrast, the agent with post-exploration (d) has extensively visited the first three rooms, while the coverage boundary has also been pushed into the final room already. This effect translates to the increased goal-reaching performance of post-exploration visible in Fig.~\ref{fig: PEworks}.

\subsection{When should the agent explore?}
Exploration and post-exploration play different roles in the IMGEP approach. Standard exploration (during goal reaching) may help improve our ability to reach a particular goal, while post-exploration especially helps to improve our ability to reach novel terrain. We therefore investigate whether both types of exploration should be jointly present, and to what extent they contribute to overall performance in our test environments. 

\begin{figure}[!b]
   \begin{minipage}{0.48\textwidth}
     \centering
     \includegraphics[width=.85\linewidth]{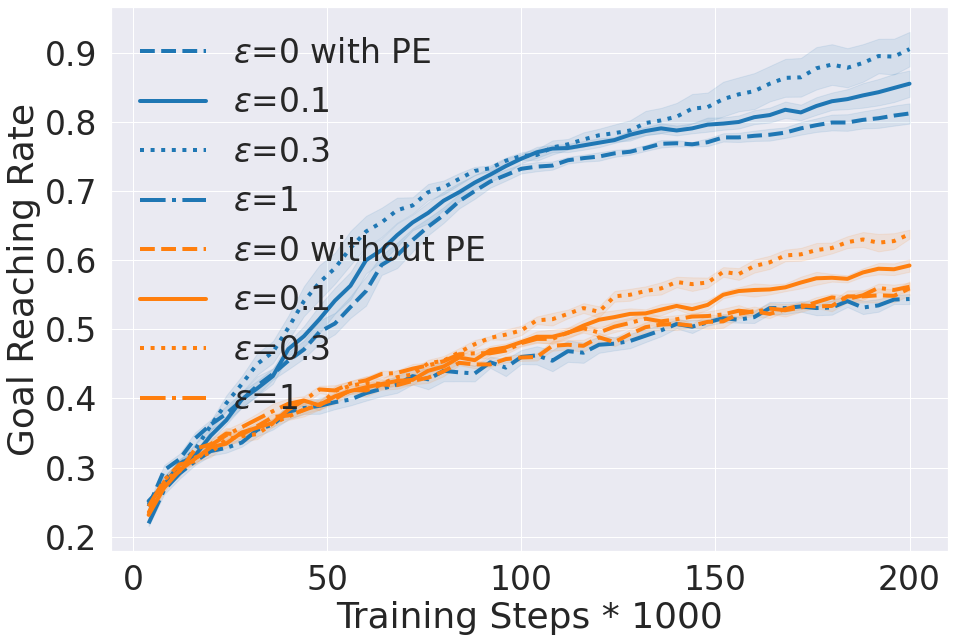}
   \end{minipage}\hfill
   \begin{minipage}{0.48\textwidth}
     \centering
     \includegraphics[width=.85\linewidth]{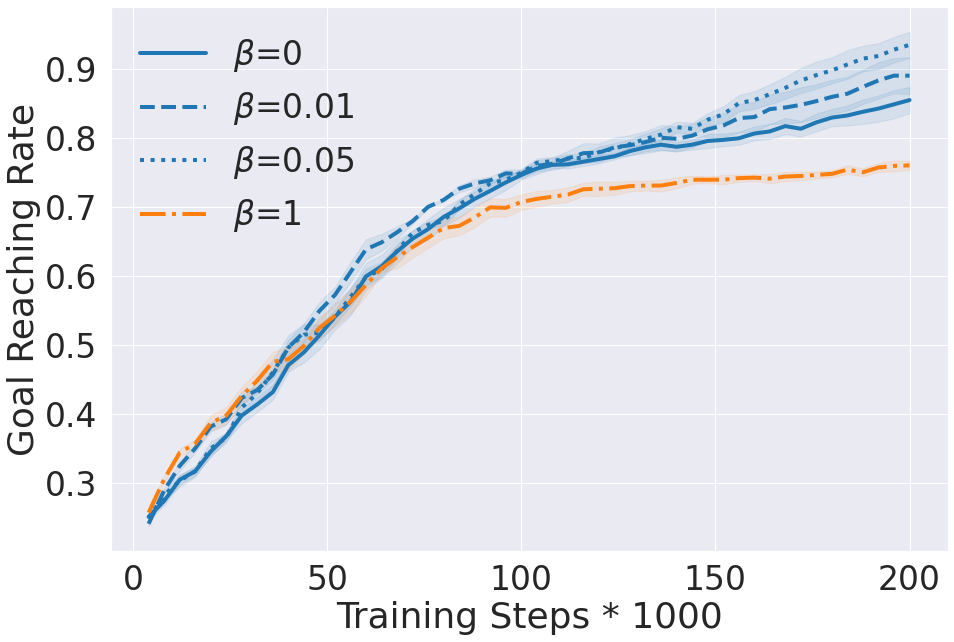}
   \end{minipage}
   \caption{{\bf Left}: Performance of agents with different exploration factors $\epsilon$ with and without Post-Exploration (PE) in the FourRooms environment. Adding slight exploration ($\epsilon=(0.1, 0.3)$) can further improve the performance. Adding fully random exploration will lead to similar results with ones without Post-Exploration. {\bf Right}: Performance of agents with different $\beta$ in the FourRooms environment. $\beta=0$ is always Post-Exploration. Selective Post-Exploration ($\beta =(0.01, 0.05)$) works better than full Post-Exploration. Too aggressive $\beta$ will hurt the performance.}
   \label{fig:eps_beta}
\end{figure}

Fig.~\ref{fig:eps_beta}, left, shows experiments on the FourRooms environments with (blue) and without (orange) post-exploration, where we also vary the amount of exploration ($\epsilon$ of an $\epsilon$-greedy policy) during goal reaching. $\epsilon=0$ implies there is no exploration at all during goal reaching, while $\epsilon=1$ implies full exploration (and therefore no exploitation at all during goal reaching).

In general, we see that post-exploration has a much stronger effect on performance than the amount of exploration during goal reaching (the quantity that is typically tuned in (goal-based) RL experiments). Only when we set $\epsilon = 1.0$ (full exploration during goal reaching) do we lose performance, which of course makes sense, since the agent then hardly manages to get back to the selected goal. Similar results on other environments are shown in Appendix.~\ref{more}.

\subsection{When should the agent post-explore?}
We next investigate when to actually post-explore, i.e., whether it is beneficial to make the probability of post-exploration dependent on the novelty of a reached goal (Eq.~\ref{eq:beta}). Results of these experiments on the FourRooms environment are shown in Fig.~\ref{fig:eps_beta}, right. For clarity, $\beta=0$ means that the agent will always post-explore, while a higher value of $\beta$ will decay the probability of post-exploration with more visits, an effect that becomes stronger the further $\beta$ increases. 

We see that selective post-exploration ($\beta=0.01$ and $\beta=0.05$) slightly outperforms full post-exploration ($\beta=0$). This selective post-exploration decides to not post-explore when it is probably not beneficial, and therefore saves it samples for a next episode. However, we should clearly not overdo it, as a faster decrease in post-exploration probability ($\beta=1$) clearly hurts performance. Note that the curve of this agent becomes almost flat near the end, probably because the agent does not post-explore anymore at all. Additional results on other environments are available in Appendix.~\ref{more}.

%


\subsection{How long should the agent post-explore?}
We next investigate whether the duration of post-exploration should be a fixed number (like in Go-Explore), or should be a function of the duration of goal-reaching (Eq. \ref{eq_n_pe}). In the former case we report a fixed $n_{pe}$ (regardless of the length of the trajectory), while in the latter case we report the used proportion $p_{pe}$. 

Fig.~\ref{fig:peANDconti}, left and middle, show the results of these experiments on FourRooms for different values of $n_{pe}$ and $p_{pe}$. The left part of the graph shows the learning curves, in which we see that $n_{pe}=20$ and $p_{pe}=0.8$ achieve best performance, suggesting that both methods are on par. However, the middle of Fig.~\ref{fig:peANDconti}, shows the total number of post-exploration steps (top) and total number of relabelling steps (bottom) in each setting. This graph shows that $p_{pe}=0.8$ uses roughly as many post-exploration steps as $n_{pe}=15$, and way less than $n_{pe}=20$. Therefore, the proportional post-exploration method ($p_{pe}$) does seem to be more effective at utilizing the post-exploration data.

\subsection{Does post-exploration work in continuing environments?}
\begin{figure}[!t]
   \begin{minipage}{0.6\textwidth}
     \centering
     \includegraphics[width=.95\linewidth]{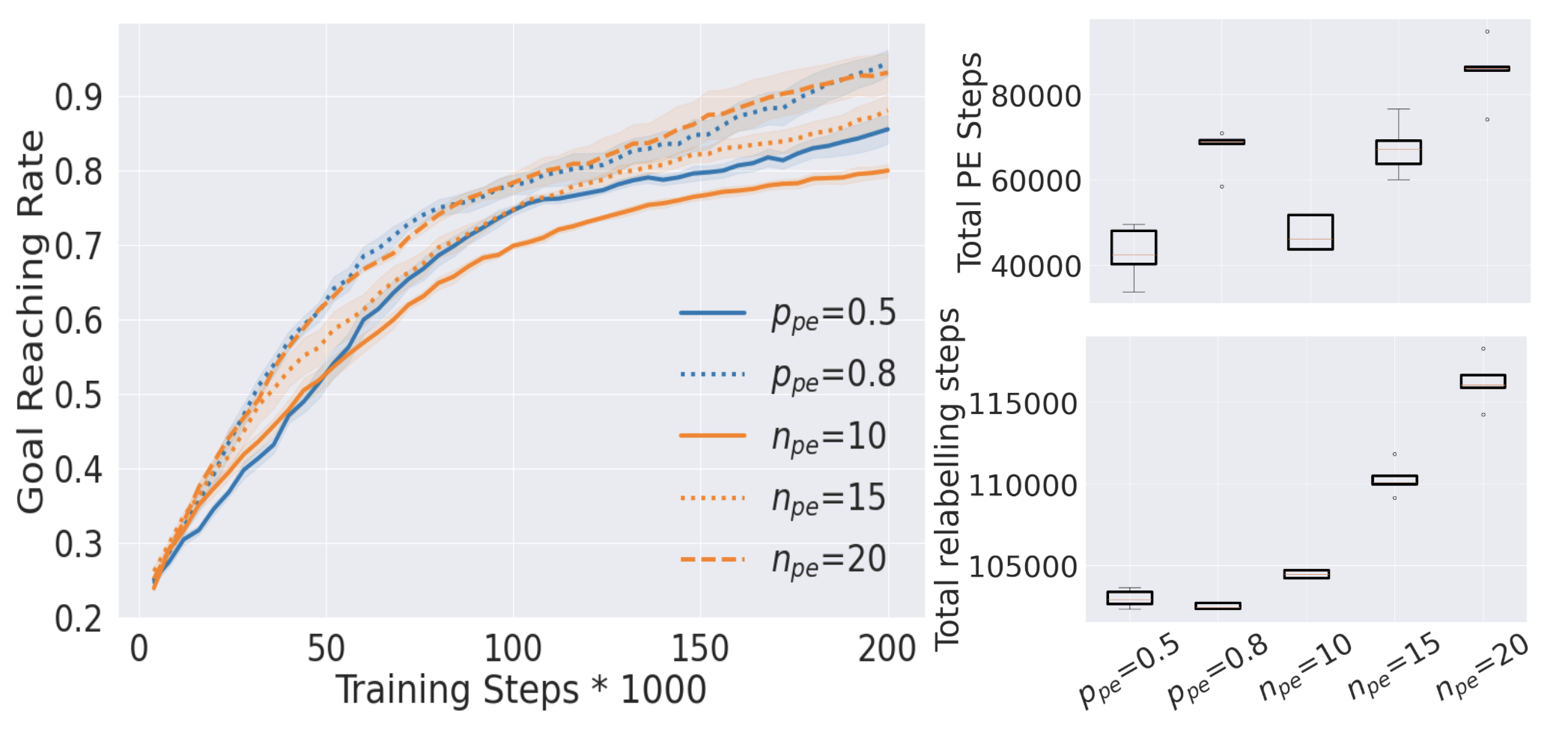}
   \end{minipage}\hfill
   \begin{minipage}{0.4\textwidth}
     \centering
     \includegraphics[width=.95\linewidth]{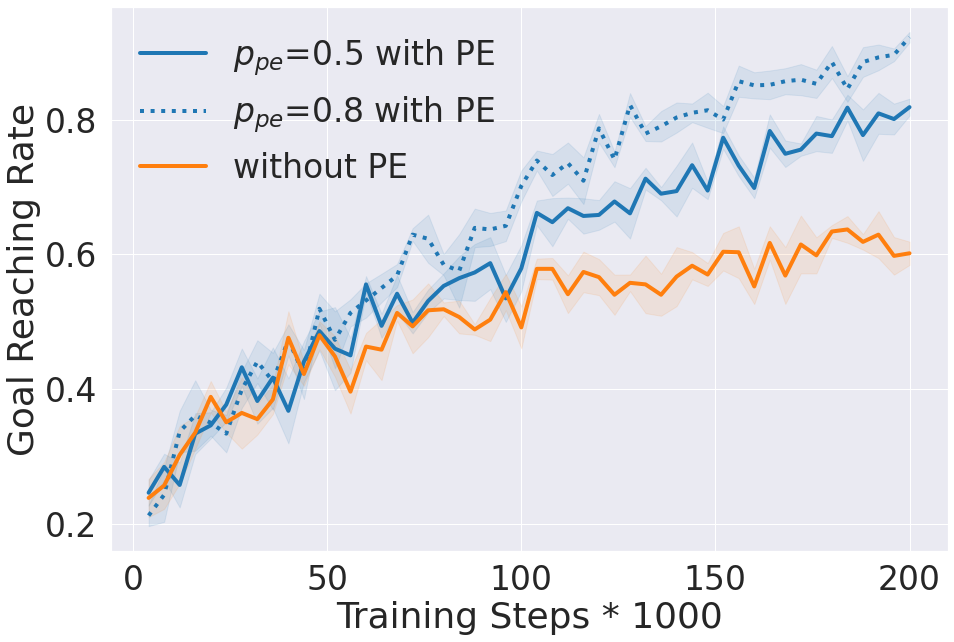}
   \end{minipage}
   \caption{{\bf Left}: Two different ways for Post-Exploration in FourRooms Environment. Left: performance of agents with different Post-Exploration methods. Blues are post-explore with percentage of the whole trajectory ($p_{pe}$) while oranges are post-explore certain steps ($n_{pe}$). {\bf Middle}: the top sub figure is total steps that agent post-explores with different Post-Exploration methods; the bottom sub figure is the total steps of hindsight relabelling. Although we keep relabelling steps and Post-Exploration steps similar for two methods, the percentage one performs better. For example, $p_{pe}=0.5$ and $n_{pe}=10$. {\bf Right}: Performance of agents with and without post-exploration in the non-episodic continuing FourRooms environment. Agents with post-exploration still outperform agents without post-exploration, likely because the next goal directs agents back into known territory.}
   \label{fig:peANDconti}
\end{figure}

Similar to Go-Explore, we investigated post-exploration primarily in the episodic setting (where we reset an agent after a goal is reached, or after post-exploration has completed). However, many specifications of goal-based RL assume a continuing task, in which there are no full resets, but the agent simply attempts to reach the next sampled goal from the current state. This does give the standard agent (without post-exploration) the ability to make additional steps after reaching a goal, but it will do so in the direction of an already set goal (which may direct it back into known territory, instead of into new terrain). Fig.~\ref{fig:peANDconti}, right, shows results of these experiments. Again, post-exploration outperforms the agent without post-exploration, although the effect is less pronounced compared to the episodic setting. This confirms our hypothesis that setting a next goal is indeed not the same as post-exploration, since the agent will likely just move back into known territory.

\section{Conclusion and Future Work}
An intrinsically motivated agent not only needs to set interesting goals and be able to reach them, but should also decide whether to continue exploration from the reached goal (`post-exploration'). In this work, we systematically investigated the benefit of post-exploration in the IMGEP framework. Experiments in several MiniGrid environments show that post-exploration is beneficial, and may even have a stronger effect on performance than tuning exploration during goal reaching (which is usually tuned in RL experiments). According to our intuition, agent with post-exploration gradually push the boundaries of their known region outwards, which allows them to reach a greater diversity of goals. Moreover, we find that {\it adaptive} post-exploration, where we adjust when and for how long we post-explore based on previous data, may further enhance this benefit. Finally, the benefit of post-exploration is retained on continuing tasks, which shows that simply setting a next goal is not the same as proper post-exploration. 

The current paper studied post-exploration in the tabular setting, to better understand its basic properties. In future work, post-exploration may be scaled up to higher-dimensional IMGEP problems which require function approximation, a learned goal space, and more complex goal sampling strategies. Moreover, our current implementation uses random post-exploration, which turned out to already work reasonably well. Another interesting direction for future work is to post-explore in a smarter way, for example by trying to set goals outside the known area, which was not possible in the tabular setting, but would be possible when we use function approximation. Altogether, post-exploration seems a promising direction for future RL exploration research. 

\clearpage

\bibliography{iclr2022_conference}
\bibliographystyle{iclr2022_conference}

\newpage
\appendix
\section{Appendix}
\label{more}
\subsection{More results on LavaCrossing}
\label{cross}
\begin{figure}[H]
    \centering
    \includegraphics[scale=0.2]{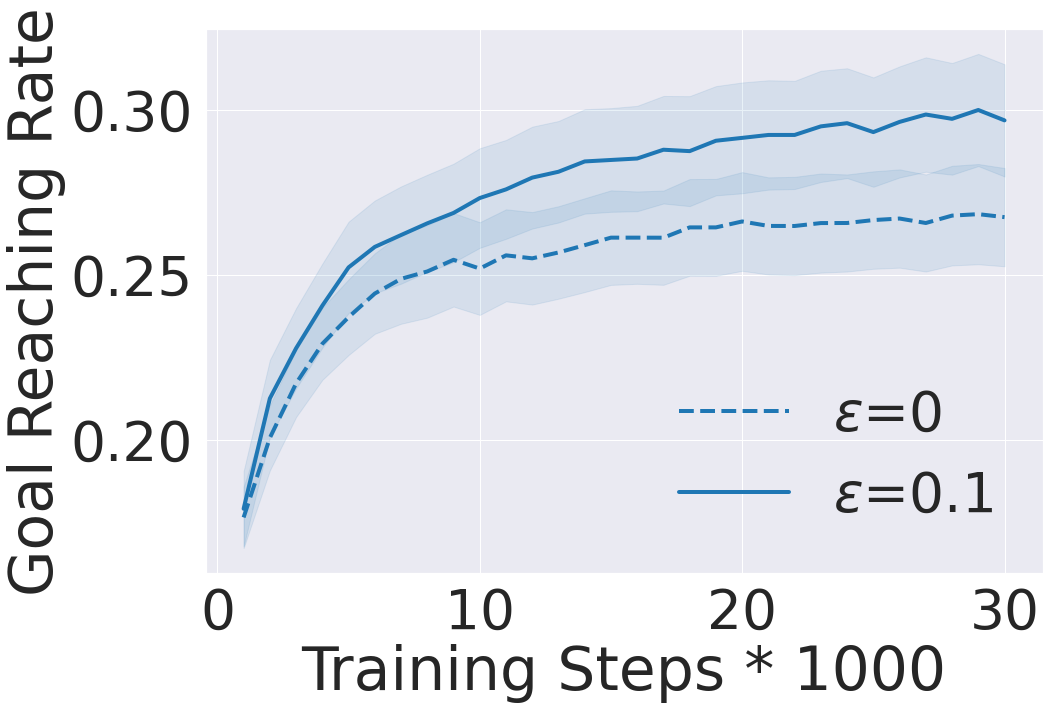}
    \caption{Performance of agents with Post-Exploration with different exploration factor $\epsilon$ in LavaCrossing. Adding exploration ($\epsilon=0.1$) during goal reaching can improve the performance. Each experiment in LavaCrossing is averaged over 10 different environment seeds and 5 repetitions.}
    \label{fig:pe_cros}
\end{figure}

\begin{figure}[!htb]
   \begin{minipage}{0.48\textwidth}
     \centering
     \includegraphics[width=.7\linewidth]{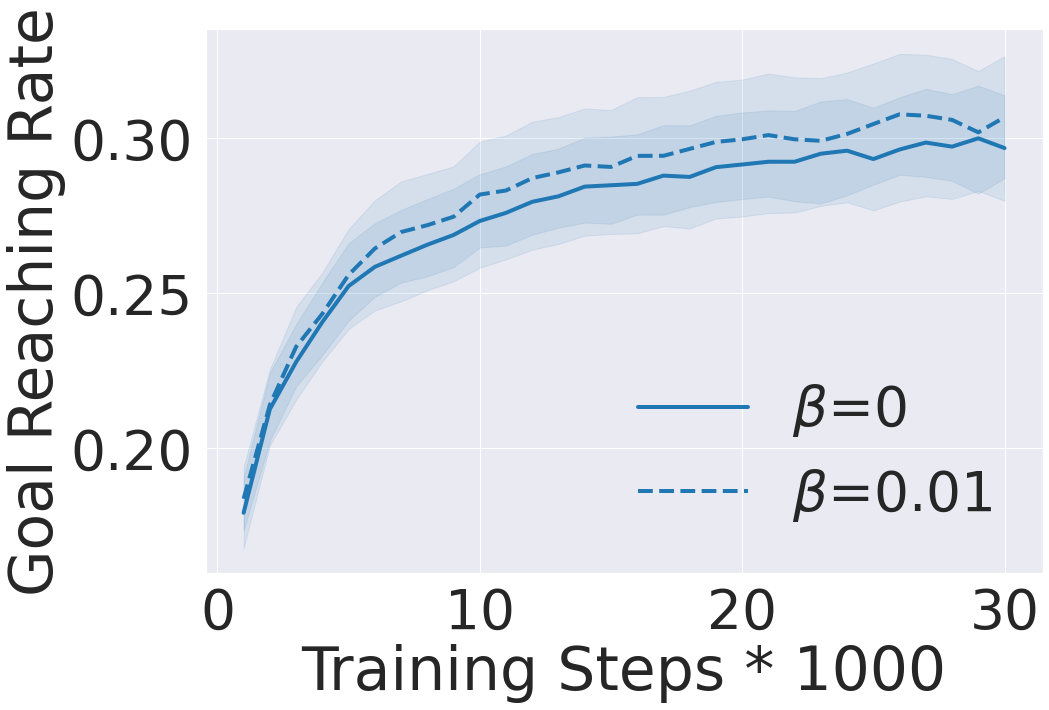}
   \end{minipage}\hfill
   \begin{minipage}{0.48\textwidth}
     \centering
     \includegraphics[width=.7\linewidth]{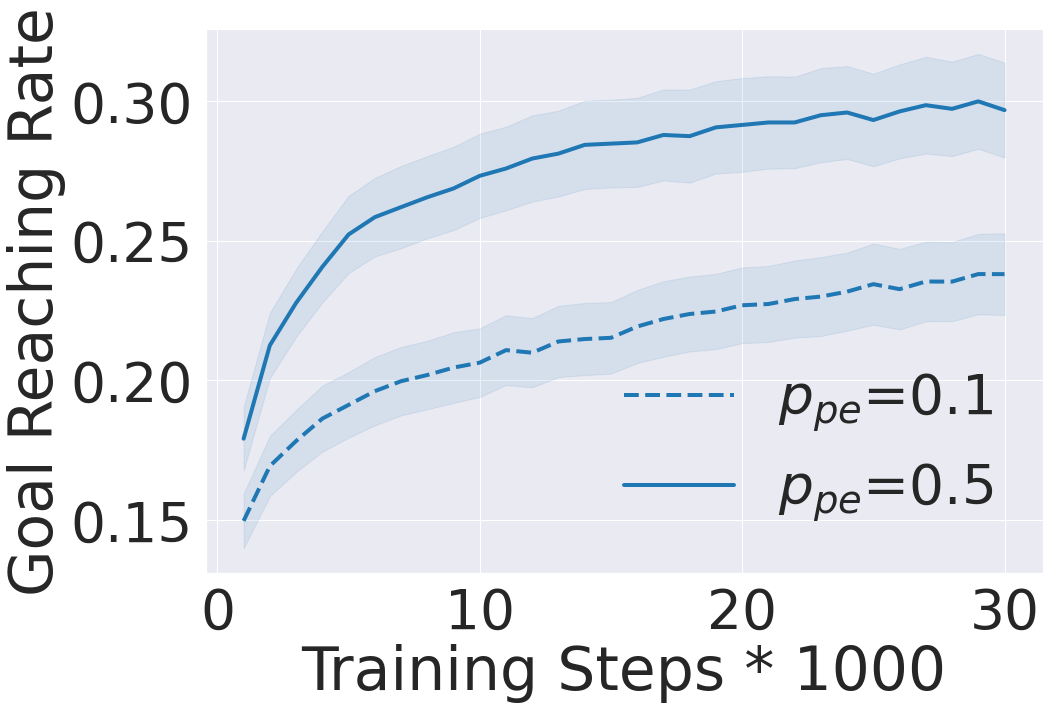}
   \end{minipage}
   \caption{Left: Performance of agents with different $\beta$. The selective Post-Exploration ($\beta=0.01$) slightly outperform full Post-Exploration ($\beta=0$). Right: Performance of agents with different $p_{pe}$. More Post-Exploration steps lead to better performance.}
   \label{fig:pe_beta_cros}
\end{figure}

\subsection{More results on LavaGap}
\label{gap}

\begin{figure}[H]
    \centering
    \includegraphics[scale=0.2]{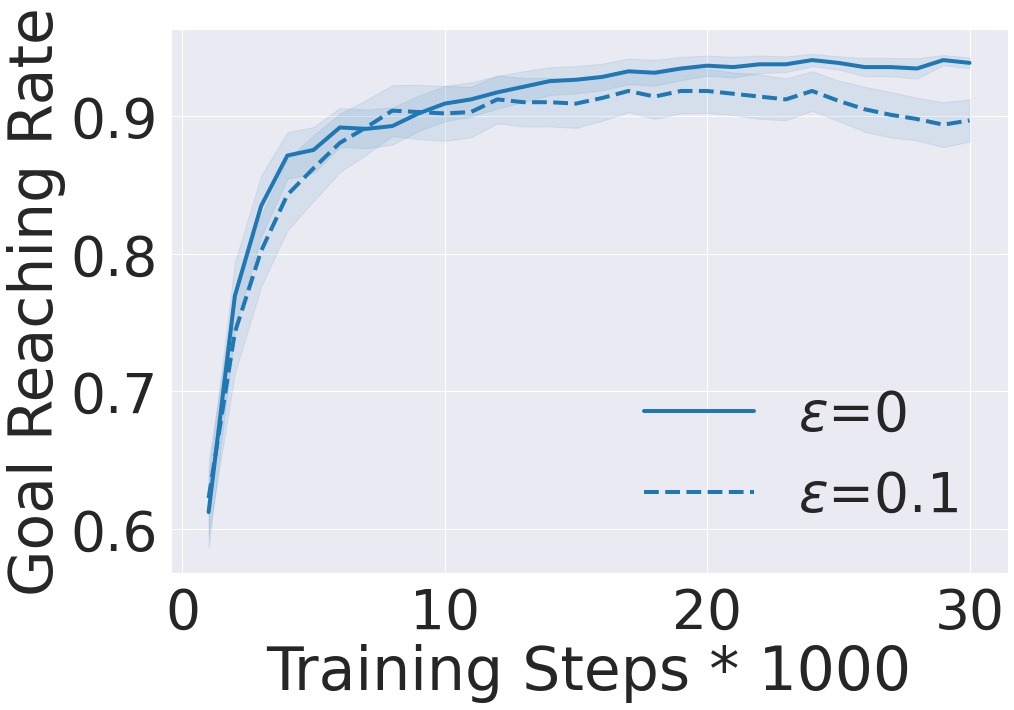}
    \caption{Performance of agents with different exploration factors $\epsilon$ in LavaGap. Adding exploration during goal reaching does not improve the overall performance. We think that's because the LavaGap environment is too small so that the exploration can barely bring the agent to new states. Each experiment in LavaCrossing is averaged over 10 different environment seeds and 5 repetitions.}
    \label{fig:pe_gap}
\end{figure}

\begin{figure}[!htb]
   \begin{minipage}{0.48\textwidth}
     \centering
     \includegraphics[width=.7\linewidth]{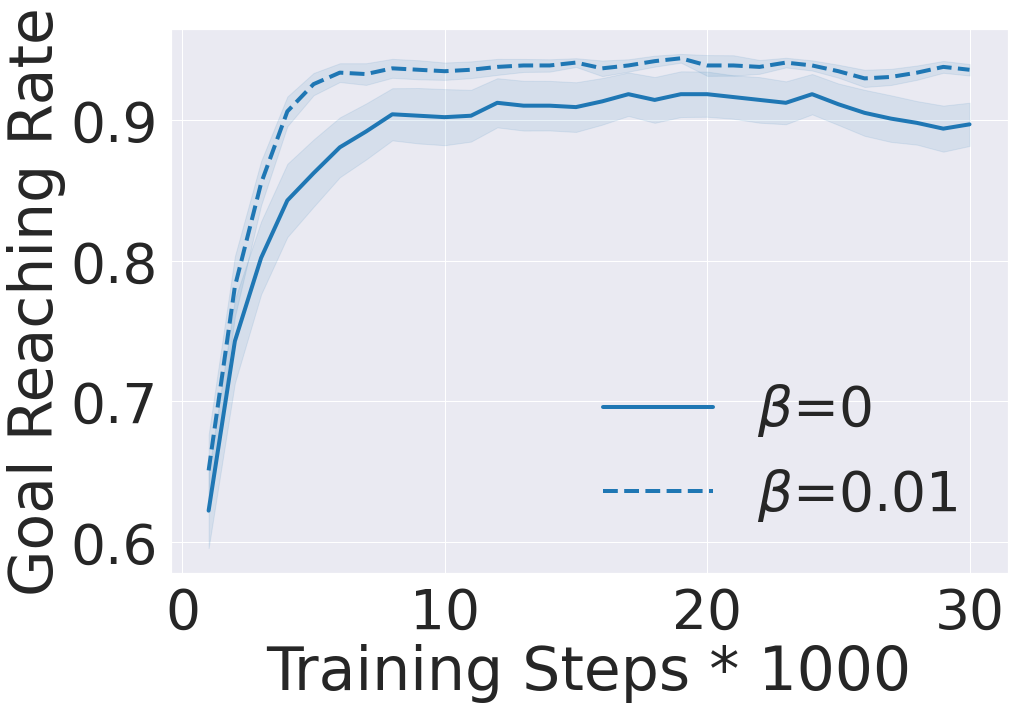}
   \end{minipage}\hfill
   \begin{minipage}{0.48\textwidth}
     \centering
     \includegraphics[width=.7\linewidth]{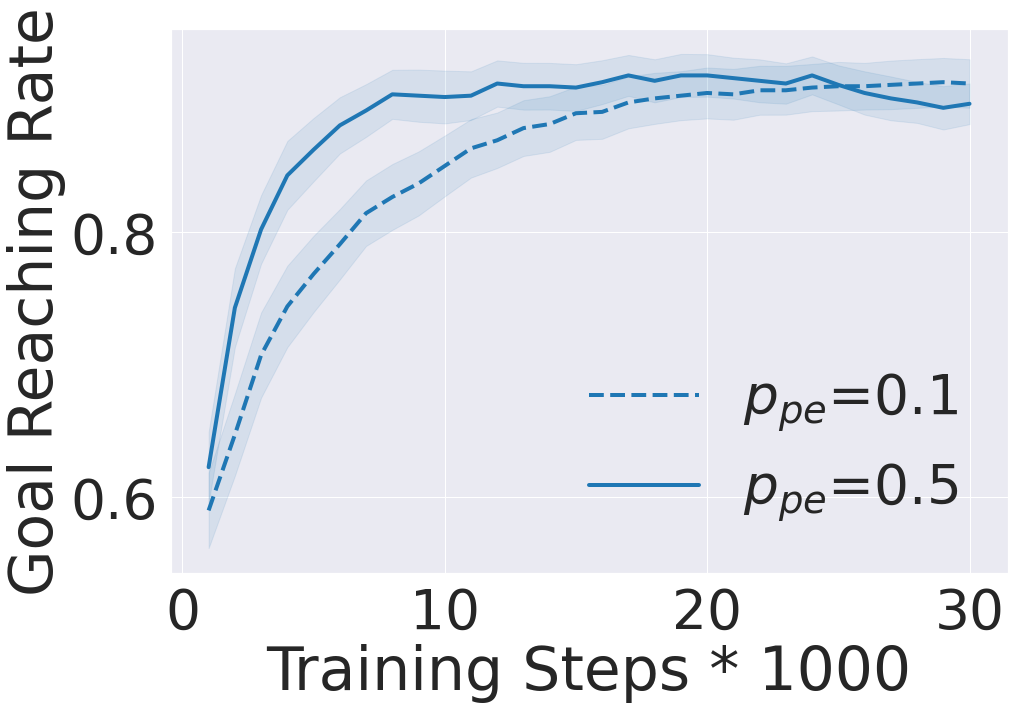}
   \end{minipage}
   \caption{Left: Performance of agents with different $\beta$. Selective Post-Exploration  ($\beta=0.01$) outperforms full Post-Exploration ($\beta=0$). Right: Performance of agents with different $p_{pe}$.}
   \label{fig:pe_beta_gap}
\end{figure}

\subsection{Hyper-parameter Settings}
\label{para}
All hyper-parameters we used in this work are shown in Tab.~\ref{tab:para}. Learning rate $\alpha$ and discount factor $\gamma$ are for Q-learning update (Eq.~\ref{eq:q}). $\beta$ is the temperature for selective Post-Exploration (Eq.~\ref{eq:beta}). $\epsilon$ is the exploration factor in $\epsilon$-greedy policy. $p_{pe}$ is the percentage of the whole trajectory that the agent will Post-Explore. $n_{pe}$ is the constant number of steps the agent will Post-Explore. Detailed explanation of each parameters can be found in corresponding sections.
\begin{table}[!htb]
    \centering
    \begin{tabular}{c|c}
         \textbf{Parameters} & \textbf{Values} \\
         \hline
        learning rate & 0.1 \\
        discount factor & 0.99 \\
        $\beta$ & 0, 0.01, 0.05, 1\\
        $\epsilon$ & 0, 0.1, 0.3, 1 \\
        $p_{pe}$ & 0.1, 0.5, 0.8 \\
        $n_{pe}$ & 10, 15, 20 \\
    \end{tabular}
    \caption{All hyper-parameters we used in this work.}
    \label{tab:para}
\end{table}

\subsection{Algorithms}
\label{alg}
Alg.~\ref{alg:pe} is the IMGEP framework we use in this work, where the red part indicates the extension with post-exploration (without the red part, the algorithm simply reduces to standard IMGEP). Alg.~\ref{alg:hr} shows how we relabel the trajectory and perform update on the relabelled data.
\vspace{1cm}

\begin{algorithm}[!htb]
   \caption{Single hindsight relabelling}
   \label{alg:hr}
\begin{algorithmic}
   \STATE {\bfseries Initialize:} Episode $M =\{g, s_0,a_0,r_0,s_1,a_1...\}$, goal-conditioned value function $Q(s,a,g)$, learning rate $\alpha \in \mathbb{R}^+$, total number of relabelled goals $k$. \\
   \STATE {\bf repeat} $k$ times: \\
   \STATE \qquad $i \gets$ set-index-to-label-as-goal() \hfill\COMMENT{avoid duplicates over repetitions} \\
   \STATE \qquad $g = s_{i}$ \hfill\COMMENT{set hindsight goal} \\
   \STATE \qquad {\bf for} $t$ in $0$ to $(i-1)$: \\
   \STATE \qquad \qquad $r_t = \mathbbm{1}_{s_t=g}$
   \STATE \qquad \qquad $\hat{Q}(s_t,a_t,g) \gets \hat{Q}(s_t,a_t,g) + \alpha \cdot[r_{t}+\gamma \cdot \max_{a} \hat{Q}(s_{t+1},a,g) - \hat{Q}(s_t,a_t,g)]$
\end{algorithmic}
\end{algorithm}

\begin{algorithm}[!htb]
   \caption{IMGEP Q-learning with post-exploration and hindsight (\textcolor{red}{red part is post-exploration})}
   \label{alg:pe}
\begin{algorithmic}
   \STATE {\bfseries Initialize:} Goal-conditioned value function $\hat{Q}(s,a,g)$, environment \texttt{Env}, episode memory $M$, goal space $G$, the number of post-exploration steps $n_{pe}$,  goal-conditioned reward function $R_g(s,a,s')$, the probability distribution of a goal $g$ being post-explored $p_{pe}(g)$, learning rate $\alpha \in \mathbb{R}^+$, number of hindsight relabels $k$. 
   \WHILE{training budget left}
   \STATE $g\sim G$ \hfill\COMMENT{sample a goal $g$ from $G$}
   \STATE $s \sim p_0(s)$ \hfill\COMMENT{reset environment}
   \STATE $ M \gets \{ \}$
   \WHILE{$s$ not terminal {\bfseries and} $g \neq s$}
   \STATE $a \gets$ \texttt{e-greedy($\hat{Q}(s,a,g)$)}
   \STATE $s' \gets$ \texttt{Env.step}($a$) \hfill\COMMENT{simulate environment}
   \STATE $r \gets R_g(s,a,s')$ \hfill\COMMENT{goal-conditioned reward}
   \STATE $\hat{Q}(s,a,g) \gets \hat{Q}(s,a,g) + \alpha \cdot[r +\gamma \cdot \max_{a'} \hat{Q}(s',a',g) - \hat{Q}(s,a,g)]$
   \STATE \texttt{$M$.append($\langle s,a,r,s',g \rangle$})
   \STATE \texttt{$G$.update($s'$)} \hfill\COMMENT{Possibly augment goal space with new observation}
   \STATE $s \gets s'$  \hfill\COMMENT{\textcolor{red}{$\Delta \sim$  Uniform[0,1]}}    
   \ENDWHILE 
   \IF{\textcolor{red}{$g$ is reached} {\bfseries and}  \textcolor{red}{$\Delta \leq p_{pe}(g)$}} 
    \FOR{\textcolor{red}{1} 
        \TO \textcolor{red}{$n_{pe}$}} 
        \STATE \textcolor{red}{$a$ = \texttt{RandomAction()}} \hfill\COMMENT{Post-exploration has random action selection}
        \STATE \textcolor{red}{$s'$ = \texttt{Env.step($a$)}}
        \STATE \textcolor{red}{$r=0, g=\texttt{None}$}
        \STATE \textcolor{red}{$M$\texttt{.append}($\langle s,a,r,s',g \rangle$)}
        \STATE \textcolor{red}{$s \gets s'$}
    \ENDFOR
   \ENDIF
   \STATE \texttt{Hindsight}($M$,$\hat{Q}(s,a,g)$, $\alpha$, $k$) \hfill\COMMENT{ Alg. \ref{alg:hr} }
\ENDWHILE
\end{algorithmic}
\end{algorithm}

\end{document}

%% file: math_commands.tex
\usepackage{amsmath,amsfonts,bm}

\def\eqref#1{equation~\ref{#1}}

\def\1{\bm{1}}

\DeclareMathAlphabet{\mathsfit}{\encodingdefault}{\sfdefault}{m}{sl}
\SetMathAlphabet{\mathsfit}{bold}{\encodingdefault}{\sfdefault}{bx}{n}

\DeclareMathOperator*{\argmax}{arg\,max}